\documentclass[letterpaper, 10 pt, conference]{ieeeconf}
\IEEEoverridecommandlockouts                          
\let\labelindent\relax
\usepackage[inline]{enumitem}
\usepackage{amsmath}
\usepackage{cite}
\usepackage{amssymb,amsfonts}
\usepackage[ruled,vlined]{algorithm2e}
\usepackage{graphicx}
\usepackage{subfloat}
\usepackage{textcomp}
\usepackage{xcolor}
\usepackage{verbatim} 
\usepackage{mathrsfs}
\usepackage{subfigure}
\usepackage{empheq}
\usepackage{courier}
\usepackage{lipsum}
\usepackage{comment}
\usepackage{mathtools, cuted}
\usepackage{bm}
\usepackage{ wasysym } 
\usepackage{soul}
\usepackage[colorlinks,citecolor=red,urlcolor=black,bookmarks=false,hypertexnames=true,hidelinks]{hyperref}

\def\BibTeX{{\rm B\kern-.05em{\sc i\kern-.025em b}\kern-.08em
    T\kern-.1667em\lower.7ex\hbox{E}\kern-.125emX}}

\newcommand{\call}{\mathcal}

\def\cA{{\call A}}
\def\cB{{\call B}}

\def\cI{{\call I}}

\def\cO{{\call O}}

\def\cX{{\call X}}
\def\cY{{\call Y}}
\def\cZ{{\call Z}}

\newcommand{\naturals}{\mathbb{N}}

\newcommand{\ap}{AP}
\newcommand{\true}{{\tt{True}}}
\newcommand{\false}{{\tt{False}}}
\newcommand{\var}{\pi}
\newcommand{\inp}{\cI}

\newcommand{\inpcen}{\inp_{\textrm{robot}}}
\newcommand{\inpterrain}{\inp_{\textrm{terrain}}}
\newcommand{\inpterrainbool}{\inp_{\textrm{terrain}}^{\cB}}

\newcommand{\inpstatesterrainexprposs}{\dom_{\textrm{terrain}}^{\textrm{poss}}}

\newcommand{\inprequest}{\inp_{\textrm{req}}}
\newcommand{\tvar}[2]{n_{#1}^{#2}}

\newcommand{\xcells}{\cX}
\newcommand{\ycells}{\cY}
\newcommand{\out}{\cO}
\newcommand{\skillcustom}[1]{\skill_{#1}}
\newcommand{\skillprecustom}[1]{\Sigma_{#1}^{\textrm{pre}}}
\newcommand{\skillpostcustom}[1]{\Sigma_{#1}^{\textrm{post}}}
\newcommand{\skill}{o}
\newcommand{\skillpre}{\skillprecustom{\skill}}
\newcommand{\skillpost}{\skillpostcustom{\skill}}

\newcommand{\dom}{\Sigma}
\newcommand{\statevar}{\sigma}
\newcommand{\inpstate}{\statevar_\inp}
\newcommand{\inpstates}{\dom_\inp}
\newcommand{\inpstateterrain}{\statevar_{\textrm{terrain}}}
\newcommand{\inpstateterrainbool}{\statevar_{\textrm{terrain}}^{\cB}}

\newcommand{\inpstatesterrain}{\dom_{\textrm{terrain}}}

\newcommand{\inpstatecen}{\statevar_{\textrm{robot}}}
\newcommand{\inpstatescen}{\dom_{\textrm{robot}}}
\newcommand{\inpstaterequest}{\statevar_{\textrm{req}}}
\newcommand{\inpstatesrequest}{\dom_{\textrm{req}}}
\newcommand{\inpstatesrequestposs}{\dom_{\textrm{req}}^{\textrm{poss}}}

\newcounter{examplecounter}
\newcommand{\examplelabel}[1]{%
  \refstepcounter{examplecounter}%
  \label{#1}%
  \theexamplecounter%
}

\newcounter{problemcounter}
\newcommand{\problemlabel}[1]{%
  \refstepcounter{problemcounter}%
  \label{#1}%
  \theproblemcounter%
}

\newcounter{defcounter}
\newcommand{\deflabel}[1]{
    \refstepcounter{defcounter}%
    \label{#1}%
    \thedefcounter%
}

\newcommand{\envspec}{\varphi_\textrm{e}}

\newcommand{\sysspec}{\varphi_\textrm{s}}
\newcommand{\envinit}{\varphi_\textrm{e}^\textrm{i}}
\newcommand{\sysinit}{\varphi_\textrm{s}^\textrm{i}}
\newcommand{\alphainit}{\varphi_\alpha^\textrm{i}}
\newcommand{\envsafety}{\varphi_\textrm{e}^\textrm{t}}

\newcommand{\alphasafety}{\varphi_\alpha^\textrm{t}}

\newcommand{\envsafetynothard}{\varphi_\textrm{e}^\textrm{t,skill}}
\newcommand{\envsafetyhard}{\varphi_\textrm{e}^\textrm{t,hard}}
\newcommand{\alphasafetynothard}{\varphi_\alpha^\textrm{t,skill}}
\newcommand{\alphasafetyhard}{\varphi_\alpha^\textrm{t,hard}}

\newcommand{\syssafety}{\varphi_\textrm{s}^\textrm{t}}
\newcommand{\syssafetynothard}{\varphi_\textrm{s}^\textrm{t,skill}}
\newcommand{\syssafetyhard}{\varphi_\textrm{s}^\textrm{t,hard}}

\newcommand{\envlive}{\varphi_\textrm{e}^\textrm{g}}
\newcommand{\syslive}{\varphi_\textrm{s}^\textrm{g}}
\newcommand{\alphalive}{\varphi_\alpha^\textrm{g}}

\newcommand{\notallowedrepair}{\varphi_\textrm{disallow}}
\newcommand{\disallowedtrans}{\notallowedrepair}

\newcommand{\spec}{\varphi}

\newcommand{\range}[1]{[#1]}

\newcommand{\strategy}{\cA}

\newcommand{\grounding}{\textrm{G}}
\newcommand{\inversegrounding}{\grounding^{-1}}
\newcommand{\ps}{\cZ}
\newcommand{\physicalstate}{z}
\newcommand{\reals}{\mathbb{R}}

\makeatother
\begin{document}

\newcommand{\vg}[1]{\bm{#1}}
\renewcommand{\v}[1]{\mathbf{#1}}

\title{\LARGE \bf Physically-Feasible Reactive Synthesis for Terrain-Adaptive Locomotion via Trajectory Optimization and Symbolic Repair\\
}

\author{
\large Ziyi Zhou*$^{1}$,
Qian Meng*$^{2}$, 
Hadas Kress-Gazit$^{2}$, and
Ye Zhao$^{1}$
\thanks{*The first two authors equally contributed to this work}
\thanks{$^{1}$Ziyi Zhou, and Ye Zhao are with Georgia Institute of Technology. {\tt\small \{zhouziyi, yzhao301\}@gatech.edu.}}
\thanks{$^{2}$Qian Meng and Hadas Kress-Gazit are with Cornell University. {\tt\small \{qm34,hadaskg\}@cornell.edu.}}
}
\maketitle
\begin{abstract}
We propose an integrated planning framework for quadrupedal locomotion over dynamically changing, unforeseen terrains.
Existing 
approaches either 
rely on heuristics for instantaneous foothold selection--compromising safety and versatility--or 
solve expensive trajectory optimization problems
with complex terrain features and long time horizons.
In contrast,
our framework leverages reactive synthesis to 
generate correct-by-construction controllers
at the symbolic level, 
and mixed-integer convex programming (MICP) 
for dynamic and physically feasible 
footstep planning 
for each symbolic transition.
We use a high-level manager to reduce the large state space
in synthesis
by incorporating local environment information, improving synthesis scalability.
To handle specifications that cannot be met due to dynamic infeasibility,
and to minimize costly
MICP solves,
we leverage a 
symbolic repair process to 
generate only necessary
symbolic transitions.
During online execution, 
re-running the MICP with real-world terrain data, along with runtime symbolic repair,
bridges the gap between offline synthesis and online execution.
We demonstrate, in simulation, 
our framework's capabilities to discover missing locomotion skills and 
react promptly
in safety-critical environments, such as scattered stepping stones and rebars.
\end{abstract}

\section{Introduction}

Terrain-adaptive locomotion is crucial for enhancing the traversing capabilities of legged robots and advancing beyond blind locomotion \cite{di2018dynamic,kim2019highly, zhou2022momentum}. Existing approaches that enable terrain-adaptive and dynamic locomotion focus on instantaneous foothold adaptation \cite{fankhauser2018robust,grandia2023perceptive} around a nominal reference trajectory. To further enhance the traversability, non-fixed gait pattern has been studied. 
In \cite{Park2015OnlinePF, kim2020vision,gilroy2021autonomous}, jumping motions are generated offline and triggered through heuristics when facing an obstacle. 
In \cite{norby2020fast,chignoli2022rapid}, the switch between normal walking and jumping is either embedded inside a reduced-order model during sampling \cite{norby2020fast} or decided by a pre-trained feasibility classifier \cite{chignoli2022rapid}. 
However, formal guarantees on locomotion safety \cite{ zhao2022reactive,shamsah2023integrated},  considering the robot's physical capabilities for traversing challenging terrains, are rarely explored—despite their importance in safety-critical scenarios like hazardous debris and construction sites.


Mixed-integer program (MIP)-based approaches \cite{valenzuela2016mixed,shirai2022simultaneous} treat both the contact state and the contact plane selection for each leg at each timestep as binary variables, and directly address the interplay between terrain segments and robot kinematics and dynamics \cite{deits2014footstep}. By relaxing the dynamics and constraints, a global certificate for the approximated convex problem (MICP) exists upon convergence, providing an ideal means to formally determine the locomotion feasibility.
However, long time horizon and complex features of surrounding terrains significantly increase the computational burden for MIP-based methods, preventing efficient online deployment.
Simplifications have been made to enhance speed, \textit{e.g.}, assuming predetermined contact sequences\cite{ding2020kinodynamic,fey20243d} and timing \cite{ponton2016convex,risbourg2022real,acosta2023bipedal}, or fixing the number of footsteps \cite{aceituno2017simultaneous}, but still face computational challenges in complicated scenarios with a large number of terrain segments.


\begin{figure}[t]
    \centering
    \includegraphics[width=\linewidth]{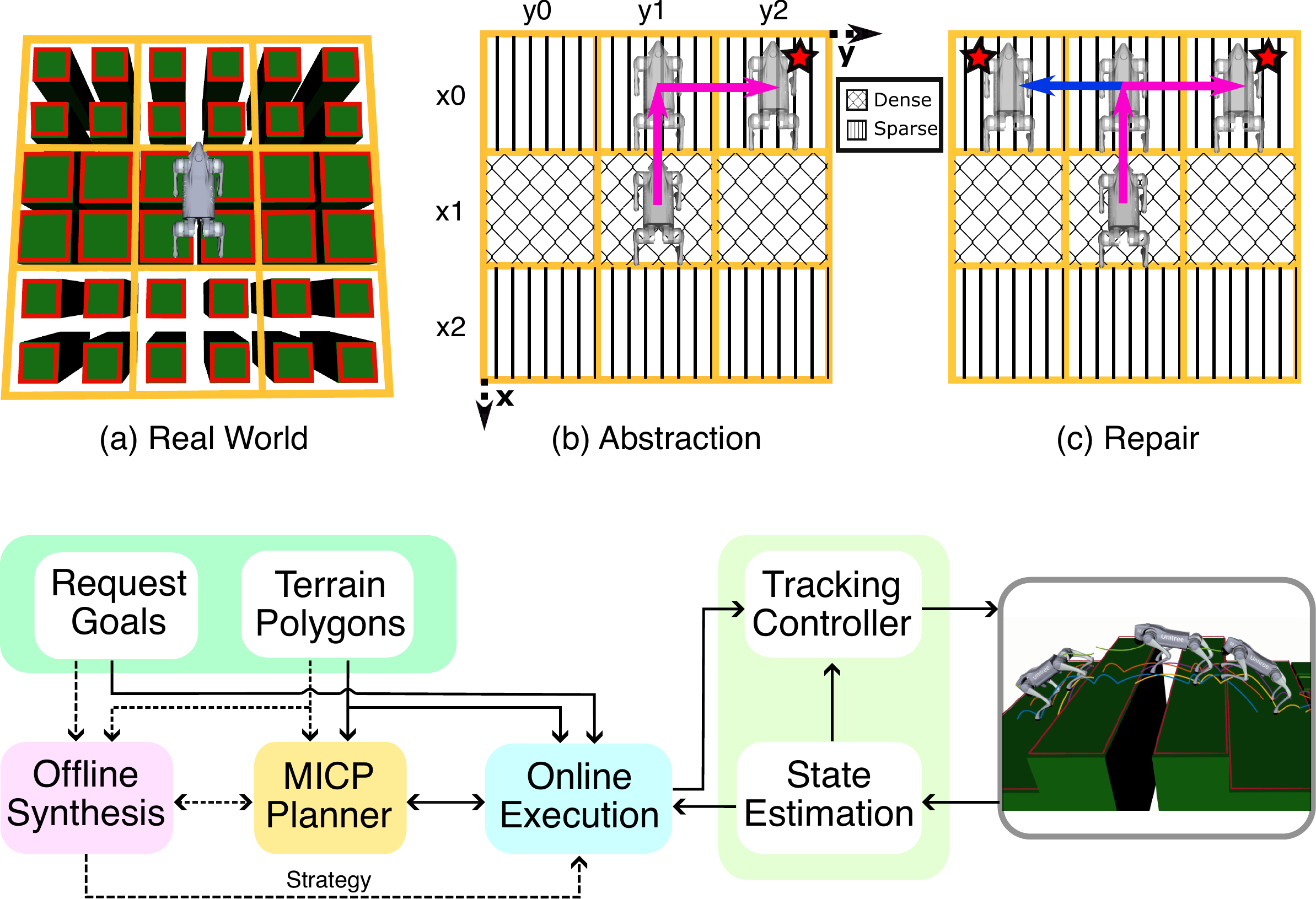}
    \caption{Upper: terrain abstraction and repair example. (a) Top-down view of the real-world terrain before abstraction. The red polygons denote the segmented terrain polygons. (b) Abstraction of the terrain and robot's skills (pink) moving from one location to another. (c) Repair process to find a new skill (blue). Lower: system architecture overview. The solid lines indicate online communication, while dashed lines represent offline processes. 
    The request goals in the offline phase are user-defined, while those in the online phase are provided by a global planner based on the global goal.}
    \label{fig:preliminary_nine_squares}
    \vspace{-0.15in}
\end{figure}


To reduce the complexity of navigating challenging terrains, one approach is to break down global, long-horizon tasks into dynamically evolving, robot-centric local environments
with intermediate waypoints \cite{fankhauser2018robust}. 
However, the corresponding MIP remains computationally expensive because it must consider all surrounding terrains. 
To address this, we discretize the local environment and restrict the MIP formulation to solving only individual discrete transitions. 
To efficiently compose these discrete transitions with formal correctness guarantees, we employ Linear Temporal Logic (LTL)-based reactive synthesis \cite{bloem2012synthesis} to generate a correct-by-construction robot strategy that guides the composition of MIP-based transitions toward the intermediate goals.

In this paper,
we
combine
reactive synthesis
and MIP-based approaches to safely and promptly react to dynamically changing environments. 
We abstract the continuous robot states and local environments 
into symbolic states, enabling high-level robot actions on the fly toward challenging terrains such as obstacles and large gaps,
as shown in Fig.~\ref{fig:preliminary_nine_squares}. 
We then solve
a MICP
to provide a physical feasibility certificate for each symbolic transition. 
This integrated framework not only bridges the gap between high-level abstraction and low-level physical capabilities but also alleviates the computational burden of MICP. 
Guided by the synthesis-based symbolic planner, 
each MICP can consider shorter time horizons and fewer terrain features than traditionally solving a single, long-horizon MICP problem.
Another noteworthy feature of our planning framework is its two-stage hierarchy. 
In the offline synthesis phase, 
we leverage
relatively expensive gait-free MICPs that optimize both contact state and foothold selection to generate the necessary locomotion gaits for symbolic transitions. 
In the online execution phase, 
we use
efficient gait-fixed MICPs with predefined
gaits to produce dynamically feasible motions and footholds on the fly.


Our offline synthesis module tackles two key scalability challenges.
First,
our task specification has a prohibitively large state space since it needs to encode the surrounding terrain states.
Since our synthesis algorithm has 
exponential time complexity in the number of variables~\cite{bloem2012synthesis},
it is inefficient and impractical to directly synthesize a controller for the full specification.
To tackle this challenge, 
we leverage an observation that the robot can continuously reason about its local environment where 
the terrains and an intermediate goal remain unchanged. 
Thus, we propose a high-level manager 
that fixes the terrain and goal information, 
only keeps the skills needed for the current terrains, 
and updates them as the robot moves in the environment. 
In our experiment,
the manager reduces the number of variables by $80.9 - 90.1\%$ depending on the number of cells and terrain types,
relieving the computation burden of synthesis. 

Our second scalability challenge
is that,
while 
the size of 
the MICP for our physical feasibility certificate is 
much smaller than the pure MIP approach,
solving gait-free MICP is still computationally expensive.
To mitigate this problem,
we leverage a symbolic repair approach~\cite{pacheck2022physically}
to
automatically suggest only the missing,
yet necessary, 
symbolic transitions,
and only solve MICP for them.
In this way, we avoid enumerating and solving the costly MICP for all possible symbolic transitions.
In our experiments, 
the symbolic repair reduces the number of expensive
MICP solves by 
$71.7 - 97.6\%$ 
depending on the terrains, compared with exhaustively iterating over all possible symbolic transitions,
ensuring that we only perform the costly gait-free MICP to generate new locomotion gaits when necessary.

Our online execution module also tackles two key challenges.
First, discrepancies in size and shape between offline-checked terrains and real-world ones 
can lead to mismatches between the desired symbolic transitions and
the robot's actual physical capabilities.
Second, encountering new terrains or 
intermediate goals
not considered in the offline phase
may also make symbolic specifications unrealizable, 
as the unforeseen scenarios are not handled during the offline phase. 
As such,
our online execution module
executes each symbolic transition
by solving an online MICP with real-world terrain data 
and prior locomotion gaits 
to generate a new nominal trajectory.
If a symbolic transition is no longer possible or the current terrains and goal were not considered offline,
we perform runtime repair,
as done in~\cite{meng2024automated},
to generate new robot skills and synthesize an updated robot strategy, allowing
continuous traversal on unexpected terrains.



The primary contributions of this paper are as follows:
\begin{itemize}
    \item We propose an integrated planning framework that combines reactive synthesis with MICP for terrain-adaptive locomotion. 
    Compared to pure MIP approaches, our method reduces computational burden via symbolic guidance and a two-stage hierarchy.
    \item During offline synthesis,
    we overcome
    scalability challenges
    by leveraging
    a high-level manager to reduce the size of the specifications based on local environment information,
    and symbolic repair to minimize the number of calls to the costly gait-free MICP,
    enabling efficient synthesis of terrain-adaptive locomotion strategies.
    \item During online execution, we address the disparity between offline synthesis and real-world terrain conditions at both the physical and symbolic levels. Our approach leverages an online MICP solver along with an online symbolic repair process to account for real-world terrain discrepancies and newly encountered conditions, enhancing 
    robustness
    and
    motion feasibility 
    at runtime.
\end{itemize}


\begin{figure*}[t]
    \centering
    \includegraphics[width=\linewidth]{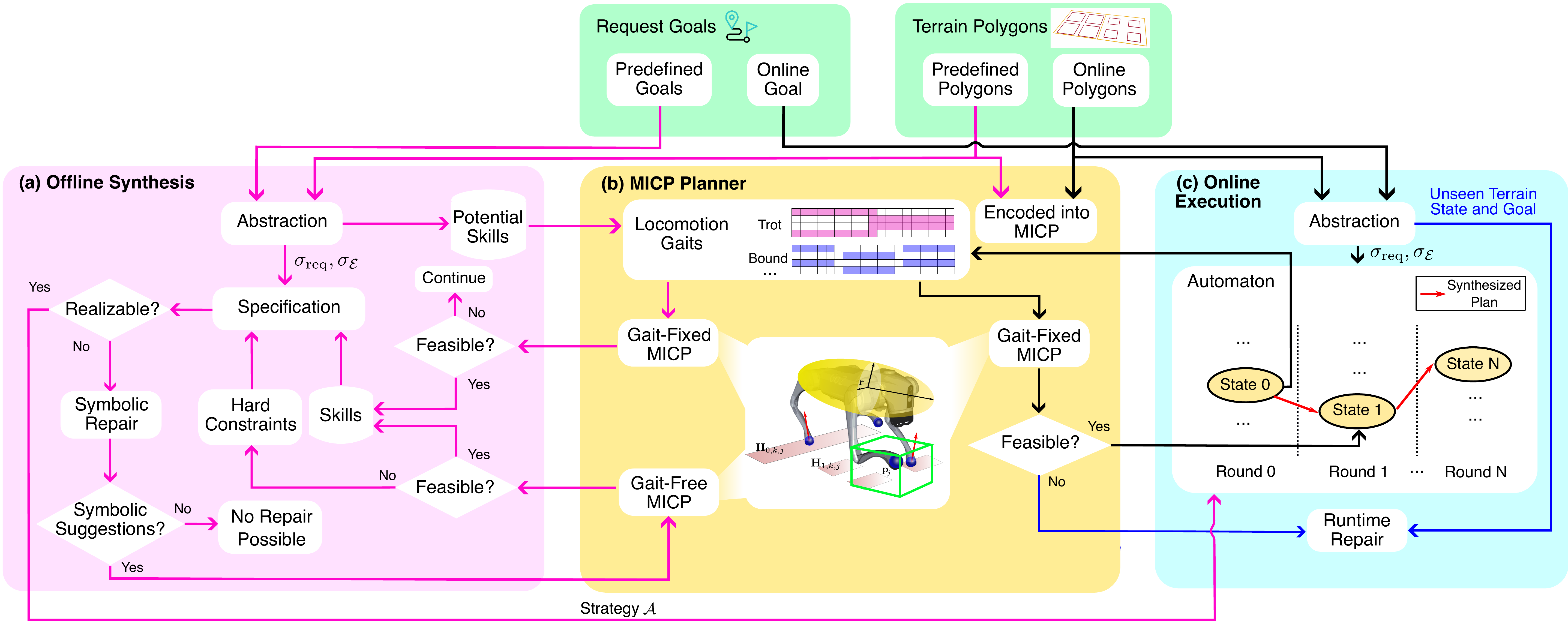}
    \caption{System overview.  
    During offline synthesis (pink arrows), an initial set of locomotion gaits is provided, and symbolic skills are iteratively generated by solving the MICP. 
    During the online execution (black arrows), MICP is solved again taking online terrain segments and the symbolic state only advances when a solution is found.
    A runtime repair (blue arrows) is initiated if a solving failure occurs, 
    or an unseen terrain or request goal is encountered.
    }
    \label{fig:combined_offline_online}
    \vspace{-0.1in}
\end{figure*}

\section{Preliminaries}
Consider a robot equipped with sensors navigating an environment toward a designated global goal $g_{\textrm{global}} \in \mathbb{R}^2$. 
This task can be decomposed into a series of local navigation tasks, where the robot moves toward intermediate, local request goals $\mathcal{G}_{\textrm{local}}$ as guided by a global planner, given surrounding segmented terrain polygons $\mathcal{P}$. 
We illustrate an example of this local task.

\noindent
\textbf{Example~\examplelabel{example:1}}.
\textit{
Consider a 2D grid world with nine cells (in yellow) in Fig.~\ref{fig:preliminary_nine_squares}. The robot is required to move from an initial cell (e.g. the center cell) 
to a desired goal cell indicated by the star. 
Each cell is assigned a terrain type, such as $\textit{Dense}$ or $\textit{Sparse}$ stepping stones.}

\subsection{Abstractions}\label{sec:preliminaries_abstraction}

Given a set of terrain polygons $\mathcal{P}$,
e.g.,
in Fig.~\ref{fig:preliminary_nine_squares}(a),
we abstract them into a 2D grid of size $n\times m$.
For each dimension,
we denote each location in that dimension with a symbol,
resulting in two sets
$\xcells \coloneqq \{x_0, \dots, x_{n-1}\}$
and
$\ycells \coloneqq \{y_0, \dots, y_{m-1}\}$.
We define a set of atomic propositions 
$\ap$,
partitioned
into sets of inputs $\mathcal{I}$
and outputs $\mathcal{O}$,
to describe the symbolic world states and robot actions.
The inputs $\mathcal{I}$ consist of
the robot inputs 
$\inpcen$,
the request inputs 
$\inprequest$,
and
the terrain inputs $\inpterrain$
($\inp = \inpcen \cup \inprequest \cup \inpterrain$).
The robot inputs 
$\inpcen \coloneqq 
\{\var_x \mid x \in \xcells\} \cup 
\{\var_y \mid y \in \ycells\}$
are used to abstract the robot position,
the request inputs
$\inprequest \coloneqq
\{\var_x^\textrm{req} \mid x \in \xcells\} \cup 
\{\var_y^\textrm{req} \mid y \in \ycells\}
$
are used to abstract the requested goal cells,
and
the terrain inputs
$\inpterrain \coloneqq \{
\tvar{x}{y} \mid x\in \xcells, y\in \ycells
\}$
describe 
the terrain type of the grid cells.
We consider a finite set of size $n_t$ of terrain types, 
defined based on their physical characteristics. 
For $\physicalstate \in \naturals$, we denote $\range{\physicalstate} \coloneqq \{0, \dots, \physicalstate - 1\}$.
We note that 
terrain inputs $\tvar{x}{y} \in \inpterrain$
are integer variables
belonging to $\range{n_t}$,
and we translate them into a set of Boolean propositions,
as done in~\cite{ehlers2016slugs}.
An input state 
$\inpstate$
is 
a value assignment function
$\inpstate: \inp \to \{\true, \false\} \cup \range{n_t}$. Since the robot can only be in one cell at a time, and the goal is a single cell, we require that exactly one $\var_x$, one $\var_y$, one $\var_x^\textrm{req}$, and one $\var_y^\textrm{req}$ be $\true$, for each input state. 
%
%
The robot states $\inpstatecen: \inpcen \to \{\true, \false\}$,
request states
$\inpstaterequest: \inprequest \to \{\true, \false\}$,
and
terrain states
$\inpstateterrain: \inpterrain \to \range{n_t}$
are the projections of $\inpstate$ on 
$\inpcen$,
$\inprequest$,
and $\inpterrain$, respectively.
We denote the sets of input states as $\inpstates$,
robot states as $\inpstatescen$,
request states as $\inpstatesrequest$,
and 
terrain states as $\inpstatesterrain$.


We use a grounding function to describe the physical meaning of 
input states.
Consider
a physical state space $\ps \subseteq \reals^n$
that represents the physical world, including the 
robot's position, orientation, and terrain properties. 
The grounding function
$\grounding: \inpstates \to 2^\ps$
maps each input state $\inpstate$ to a set of physical states in $\ps$.
For robot states
$\inpstatecen \in \inpstatescen$,
we define 
$\grounding(\inpstatecen) \coloneqq \{
\physicalstate \in \ps \mid
\exists \var_x, \var_y \in \inpcen. 
\inpstatecen(\var_x) \land \inpstatecen(\var_y) \land
\textrm{robot\_pose}(\physicalstate) \in \textrm{cell}(x,y)
\}$.
Similarly, for request states 
$\inpstaterequest \in \inpstatesrequest$,
we define
$\grounding(\inpstaterequest) \coloneqq \{
\physicalstate \in \ps \mid
\exists \var_x^\textrm{req}, \var_y^\textrm{req} \in \inprequest. 
\inpstaterequest(\var_x^\textrm{req}) \land \inpstaterequest(\var_y^\textrm{req}) \land
\textrm{request\_goal}(\physicalstate) \in \textrm{cell}(x,y)
\}$.
For terrain states $\inpstateterrain \in \inpstatesterrain$,
we define
$\grounding(\inpstateterrain)$
to be the set of physical states
whose abstracted terrain types are indicated by $\inpstateterrain$.
%
For an input state $\inpstate \in\inpstates$ whose
projections are $\inpstatecen$, $\inpstaterequest$, and $\inpstateterrain$,
$\grounding(\inpstate) \coloneqq \grounding(\inpstatecen) \cap \grounding(\inpstaterequest) \cap \grounding(\inpstateterrain)$.
%
%
Lastly,
we use an inverse grounding function 
$\inversegrounding: \ps \to \inpstates$
to map a physical state $\physicalstate \in \ps$
to its corresponding input state.

In Example~\ref{example:1},
the inputs are
$\inpcen \coloneqq 
\{
\var_{x_0}, \var_{x_1}, \var_{x_2},
\var_{y_0}, \var_{y_1}, \var_{y_2}
\}$,
$\inprequest \coloneqq \{\var^\textrm{req} \mid \var \in \inpcen\}$,
and
$\inpterrain \coloneqq \{\tvar{x_i}{y_j} \mid i,j\in 
\{0,1,2\}
\}$
where 
$\tvar{x_i}{y_j} = 1$ if the terrain type of the grid cell $(x_i, y_j)$ is Dense,
and $\tvar{x_i}{y_j} = 0$ if it is Sparse.
The input state in Fig.~\ref{fig:preliminary_nine_squares}(b) is
$\inpstate: 
\var_{x_1} \mapsto \true, 
\var_{y_1} \mapsto \true,
\var_{x_0}^\textrm{req} \mapsto \true, 
\var_{y_2}^\textrm{req} \mapsto \true,
\tvar{x_1}{y_j} \mapsto 1$ for $j\in\{0,1,2\}$.
In this paper,
we omit the Boolean propositions that are $\false$ and integer propositions
whose values are $0$ in the input states for space.

The outputs
$\out$ 
represent
the skills 
that allow the robot to transit among grid cells,
given their corresponding terrain states.
A skill $\skill \in \out$
consists of 
sets of preconditions $\skillpre \subseteq \inpstatescen \times \inpstatesterrain$,
from which the skill is allowed to execute,
and 
postconditions $\skillpost \subseteq \inpstatescen$, the resulting grid cell after executing the skill.
In Example~\ref{example:1},
the skill $\skillcustom{0}$ moves the robot from the middle to the upper terrain grid cell (Fig.~\ref{fig:preliminary_nine_squares}b).
The precondition of $\skillcustom{0}$ is
$\skillprecustom{\skillcustom{0}} = \{
(\inpstatecen, \inpstateterrain): \var_{x_1}\mapsto \true, 
\var_{y_1} \mapsto \true,
\tvar{x_1}{y_0} \mapsto 1, 
\tvar{x_1}{y_1} \mapsto 1, 
\tvar{x_1}{y_2} \mapsto 1
\}$, 
The postcondition is
$\skillpostcustom{\skillcustom{0}} = \{
\inpstatecen:
\var_{x_0} \mapsto \true, 
\var_{y_1} \mapsto \true
\}$.
In addition,
each skill consists of
a continuous-level locomotion gait, 
e.g. a one-second trotting gait $L$,
that physically implements the symbolic transition
(Sec.~\ref{sec:motion_primitive}).


\subsection{Specifications}
We use the Generalized Reactivity(1) (GR(1)) fragment of linear temporal logic (LTL)~\cite{bloem2012synthesis}
to encode task specifications.
LTL specifications $\spec$ 
follow the grammar 
$\spec \: := \var \:|\: \neg\: \spec \:|\: \spec \:\land\: \spec \:|\: \bigcirc\spec \:|\: \square \spec \:|\: \lozenge \spec$,
where $\var \in \ap$ is an atomic proposition,
$\neg$ ``not'' and $\land$ ``and'' are Boolean operators,
and
$\bigcirc$ ``next'',
$\square$ ``always'',
and $\lozenge$ ``eventually''
are temporal operators.
We refer the readers to~\cite{baier2008principles} for a detailed description of LTL.

GR(1) specifications are expressed in the form of 
$\spec = \envspec \to \sysspec$,
where $\envspec = \envinit \land \envsafety \land \envlive$ 
is the assumptions on the behaviors of the possibly adversarial environment,
and $\sysspec = \sysinit \land \syssafety \land \syslive$
represents the guarantee of the desired robot's behaviors.
For $\alpha \in \{e, s\}$,
$\alphainit, \alphasafety, \alphalive$
characterize the 
initial conditions,
safety constraints,
and liveness conditions,
respectively.
For symbolic repair (see Sec.~\ref{sec:manager_and_repair}),
we divide safety constraints ($\envsafety, \syssafety$) into
skill constraints ($\envsafetynothard, \syssafetynothard$)
and hard constraints ($\envsafetyhard, \syssafetyhard$),
i.e., for $\alpha \in \{e, s\}$,
$\alphasafety = \alphasafetynothard \land \alphasafetyhard$.
The skill constraints encode the pre and postcondition of skills 
(see Sec.~\ref{sec:spec_gen})
and can be modified by repair,
while repair cannot modify the hard constraints.



In practice, the GR(1) specifications are large in size due to the necessity of 
encoding terrain and task information.
Since the exact current terrain and request state information is available at runtime,
we can generate a shorter and more efficient version of the specification
via \textit{partial evaluation},
which substitutes a subset of the Boolean propositions
with their truth values to reduce the state space.

\noindent\textbf{Definition~\deflabel{def:partial_eval}.}
Given an LTL formula $\spec$
and
two subsets $S^\true, S^\false \subseteq \ap$, $S^\true \cap S^\false = \emptyset$,
we define
the \textit{partial evaluation} of $\spec$ over $S^\true, S^\false$,
as $\spec[S^\true, S^\false]$, 
where we substitute 
propositions $\pi\in AP$ in $\spec$ with $\true$ if $\pi \in S^\true$ and with $\false$ if $\pi \in S^\false$.

\section{Problem Statement}\label{sec:problem_statement}



\textbf{Problem~\problemlabel{problem:1}}.
Given 
\begin{enumerate*}[label=(\roman*),ref=\roman*]
    \item a global goal $g_{\textrm{global}}$, 
    \item a set of possible local request goals $\mathcal{G}_{\textrm{local}}$
    \item a set of predefined terrain polygons $\mathcal{P}_{\text{pre}}$ from user's prior knowledge,
    and
    \item a set of online terrain polygons $\mathcal{P}_{\text{on}}$ perceived during execution that may differ from the predefined ones,
    \item a set of predefined locomotion gaits $\mathcal{L}$;
\end{enumerate*}
generate controls that enable the robot to navigate the environment and reach the goal.


\section{Approach Summary}
To efficiently solve Problem~\ref{problem:1}, 
we manage complexity 
at both symbolic and physical levels. 
At the symbolic level, 
we leverage reactive synthesis to 
decompose the local navigation problem 
into manageable subproblems
whose solutions can be reused by synthesis
for different scenarios.
Each subproblem corresponds to finding controls for a short-horizon symbolic transition
and is solved via a MICP to 
provide physical feasibility certificates of the transition.
Our framework consists of offline synthesis (Sec.\ref{sec:reactive_gait_synthesis}) and online execution (Sec.\ref{sec:online_execution}).
During the offline phase, we generate a set of potentially useful skills and corresponding locomotion gaits based on predefined terrain states and goals, 
while leveraging symbolic repair to 
find missing yet necessary symbolic transitions.
At runtime, 
we leverage runtime repair to identify new symbolic transitions necessary for unforeseen terrain configurations or intermediate request goals.


\section{Offline Synthesis}\label{sec:reactive_gait_synthesis}
The offline synthesis module 
generates a strategy 
that enables the robot
to reach local request goals over predefined terrain states.
As shown in Fig.~\ref{fig:combined_offline_online}(a),
this module takes in
a set of local request goals $\mathcal{G}_{\textrm{local}}$ and
a set of possible terrain polygons $\mathcal{P}_{\text{pre}}$ from prior knowledge of the workspace. 
In addition, the user provides a set of locomotion gaits $\mathcal{L}$. 
We first 
leverage the inverse grounding function $\inversegrounding$ to 
discretize the local environment,
obtain
a set of possible
request states $\inpstatesrequestposs \subseteq \inpstatesrequest$
from the local request goals,
and characterize the predefined set of terrain polygons into a set of possible terrain states $\inpstatesterrainexprposs \subseteq \inpstatesterrain$.
%
Next, 
we solve a gait-fixed MICP problem to determine the feasibility of each potential skill given the locomotion gaits
(Sec.~\ref{sec:motion_primitive}), and then encode all feasible transitions as skills in a specification
(Sec.~\ref{sec:spec_gen}).
Next,
we use a high-level manager to 
generate partial evaluations of the encoded specifications
for efficient synthesis.
If any partial evaluation is unrealizable, 
we use a repair process to suggest new robot skills 
and a gait-free MICP to check the feasibility of the suggested skills, 
eventually making the specification realizable
(Sec.~\ref{sec:manager_and_repair}).



\subsection{Locomotion Gait and Feasibility Checking via MICP}\label{sec:motion_primitive} 
We assume each locomotion gait $L$ comes with a contact sequence $\mathcal{G}$ and corresponding time durations $T$. Therefore, we define the locomotion gait as $L = \mathcal{M}(\mathcal{G}, T)$. Each skill is assumed to have a unique locomotion gait to determine how the robot moves at the continuous level. 
As shown in Fig.~\ref{fig:combined_offline_online}(a), since all the possible request states $\inpstatesrequestposs$ and terrain states $\inpstatesterrainexprposs$ are given during the offline phase, it is straightforward to first identify all the potential skills at the symbolic level that allow the robots to move freely in all possible local environments.  Each of them is evaluated by a gait-fixed MICP using the provided locomotion gaits.

\begin{figure}[t]
    \centering
    \begin{minipage}{0.9 
    \linewidth}
    \footnotesize
    \begin{subequations}\label{ocp:mip_general}
    \begin{alignat}{2}
        \nonumber &\underset{\vg \phi, \v H_{r,k,j}}{\min} \hspace{0.1cm} &&\sum_{i = 0}^{N-1} \hspace{2pt} \delta \vg \phi[i]^T\mathbf{Q}\hspace{2pt}\delta \vg \phi[i] + \vg \phi[i]^{T}\mathbf{R}\vg \phi[i]\\
        \nonumber
        \\\label{eq:dynamics}
        &(\textit{\rm Dynamics}) &&  
        \begin{bmatrix} \v r[i+1]\\\
        \dot{\v r}[i+1]\\\
        m \ddot{\v r}[i]\\\
        \vg \theta[i+1]\\
        \dot{\vg \theta}[i+1]
        \end{bmatrix}= \begin{bmatrix}
        \v r[i] + \Delta t \cdot \dot{\v r}[i+1]\\
        \dot {\v r}[i] + \Delta t \cdot \ddot{\v r}[i+1]\\
        \sum\nolimits_{j} \v f_j[i] + m \v g \\
            \vg \theta[i] + \Delta t \cdot \dot{\vg \theta}[i+1]\\
            \dot{\vg \theta}[i] + \Delta t \cdot \ddot{\vg \theta}[i+1]
	    \end{bmatrix}
        \\\nonumber
        &(\textit{\rm Foothold}) &&\forall i \in \mathcal{C}_{k,j}[0], r \in [R], k \in [n_s], j \in [n_f]
        \\\label{eq:safe_region}
        & && \v H_{r,k,j} \Rightarrow \v A_r \v p_j[i] \leq \v b_r
        \\\nonumber & && \hspace{1.3cm} \v A_{\text{eq},r} \v p_j[i] = \v b_{\text{eq},r}
        \\\nonumber
        & &&\sum_{r=0}^{R-1}\v H_{r,k,j} = 1
        \\\nonumber
        & &&\v H_{r,k,j} \in \{0, 1\}
        \\\label{eq:force_non_zero}
        &(\textit{\rm Frictional}) 
        &&\v f_j[i] \cdot \v n(\v p_j^{xy}[i]) \ge 0, \ \forall i \in \mathcal{C}_j
        \\\label{eq:force_friction_cone}
        & &&\v f_j[i] \in \mathcal{F}(\mu, \v n, \v p_j^{xy}[i]), \ \forall i \in \mathcal{C}_j
         \\\label{eq:ee_velocity_zero}
            &(\textit{\rm Contact}) 
            &&\dot{\v p}_j[i]=0, \ \forall i \in \mathcal{C}_j
        \\\label{eq:force_zero}
         & && \v f_j[i] = 0, \ \forall i \notin \mathcal{C}_j
         \\\label{eq:max_torque}
         &(\textit{\rm Actuation})
         && \v J_j^{\top} \v f_j[i] \leq \vg \tau_{\rm max}, \forall i \in [N]
         \\\label{eq:max_acc}& && \v I \ddot{\vg \theta}[i] \leq \vg \tau_{\rm max}^{\prime}, \forall i \in [N]
        \\&(\textit{\rm Kinematics}) && \forall i \in [N], j \in [n_j]
        \\\label{eq:rom}
            & &&\v p_j[i] \in \mathcal{R}_j(^\mathcal{B} \v p_j^{\text{ref}}, \v r[i],\vg \theta_{\text{ref}}, \v p_j^{\text{max}})
    \end{alignat}
\end{subequations}
\end{minipage}
    \caption{MICP formulation for physical feasibility checking.}
    \label{fig:micp}
    \vspace{-0.2in}
\end{figure}

For the MICP formulation shown in Fig.~\ref{fig:micp}, the decision variables $\vg \phi$ include base position $\v r$, velocity $\dot{\v r}$, acceleration $\ddot{\v r}$, orientation $\vg \theta$ parameterized by Euler angle and its first and second-order derivatives $\dot{\vg \theta},\ddot{\vg \theta}$, individual end-effector (EE) position $\v p_j$, velocity $\dot{\v p}_j$, and acceleration $\ddot{\v p}_j$, and individual contact force $\v f_j$ for the foot $j$.
We define binary variables $\v H_{r,k,j}$, with $r$, $k$, and $j$ expressing the $r^{\rm th}$ convex region, $k^{\rm th}$ footstep, and $j^{\rm th}$ foot. 
We use $N$, $n_f$, $n_s$, and $R$ to represent the number of timesteps, the number of feet, the number of footsteps specified by the gait configuration for each foot, and the number of convex terrain polygons to be considered, respectively. For example,  Fig.~\ref{fig:combined_offline_online}(b) shows a case with four polygons.
The cost function consists of tracking costs and regularization terms governed by diagonal matrices $\v Q$ and $\v R$.
The tracking costs include the deviation from a desired base trajectory interpolating a path between the terrain grid locations defined in each skill.


We encode the system dynamics as a simplified single rigid body in Eq.~\ref{eq:dynamics} to keep the dynamics constraint convex. To incorporate safe region constraints to select proper footholds,
Eq.~\ref{eq:safe_region} restricts
the robot's EE to stay within one of the convex polygons. We use $\mathcal{C}_{k,j}$ to denote the set of time steps indicating stance after the $k^{\rm th}$ footstep for the $j^{\rm th}$ foot and the safe region constraint only applies to the first stance time step represented as $\mathcal{C}_{k,j}[0]$.
Similarly, the collision avoidance constraint keeps each EE within a collision-free box using additional binary variables, omitting details for brevity.
Frictional constraints are defined in Eqs.~\ref{eq:force_non_zero} - \ref{eq:force_friction_cone}, with $\v n$ as the normal vector of the terrain corresponding to the position $\v p^{xy}_j$ and $\mathcal{C}_j$ as the set of timesteps indicating stance for the $j^{\rm th}$ foot. Contact constraints in Eqs.~\ref{eq:ee_velocity_zero} - \ref{eq:force_zero} check the given contact states and activate at different timesteps.
Since the joint angles are omitted in the single rigid body model, we use a fixed Jacobian $\v J_j(\v q_j^{\text{ref}})$ at a nominal joint pose $\v q_j^{\text{ref}}$ for each leg to approximately consider the torque limit constraint in Eq.~\ref{eq:max_torque}. In addition, since the translational and angular motions are decoupled, another actuation constraint on the angular acceleration is also added in Eq.~\ref{eq:max_acc}, where $\vg \tau_{\rm max}$ is the joint torque limit, $\vg \tau_{\rm max}^{\prime}$ is the torque limit applied on the base, and $\v I$ is the moment of inertia for the approximated single rigid body.
Lastly, the kinematics constraint in Eq.~\ref{eq:rom} strictly limits the possible EE movements to assure safety, where $\mathcal{R}_j$ is defined as a 3D box constraint around a nominal foot EE position $^\mathcal{B} \v p_j^{\text{ref}}$ based on the base position and reference orientation $\vg \theta_{\text{ref}}$, and constrained by a maximum deviation $\v p_j^{\text{max}}$.
Due to the convex nature of this MICP, we can determine the feasibility of a possible symbolic transition by checking if the optimal solution exists.

\subsection{Task Specification Encoding}\label{sec:spec_gen}

After acquiring a set of feasible robot skills, 
we encode the skills into the specification $\spec$ 
as part of the environment safety assumptions $\envsafety$ 
and system safety guarantees $\syssafety$.
The skill assumptions $\envsafetynothard$ encode the postconditions of the skills. 
The assumptions ensure that after the skill execution, 
at least one postcondition of the skill must hold:
$\envsafetynothard \coloneqq \bigwedge_{\skill \in \mathcal{O}} 
\square
( \skill 
    \to
    \bigvee_{\statevar\in\skillpost} 
    \bigwedge_{\var \in \inpcen}
    {\bigcirc (\var=\statevar(\var))} 
    )
$.
%
In Example~\ref{fig:preliminary_nine_squares},
the postcondition of skill $\skillcustom{0}$
is defined as
$
\square (\skillcustom{0} 
    \to \bigcirc \var_{x_0} \land \bigcirc \var_{y_1} \land \neg \dots)
$.

The skill guarantees $\syssafetynothard$ encodes the preconditions of the skills.
The guarantees impose restrictions on when a skill can be executed.
The guarantees only allow the robot to execute a skill 
if one of the skill's preconditions holds:
$
\syssafetynothard \coloneqq 
    \bigwedge_{\skill \in \mathcal{O}} 
    \square (
    \neg (
    \bigvee_{\statevar\in\skillprecustom{\skill}}
    \bigwedge_{\var\in \inpcen \cup \inpterrain}
    {\bigcirc(\var = \statevar(\var))}
    )
    \to \neg \bigcirc \skill)
$.
%
In Example~\ref{example:1}, the precondition of skill $\skillcustom{0}$ is defined as
$
\square \big( \neg 
     (
     \bigcirc x_1 \land 
     \bigcirc y_1 \land 
     \bigcirc \tvar{x_1}{y_0}=1 \land 
     \bigcirc \tvar{x_1}{y_1}=1 \land 
     \bigcirc \tvar{x_1}{y_2}=1 \land 
     \dots)
     \to \neg \bigcirc \skillcustom{0} \big).
$

The hard constraints $\envsafetyhard$ and $\syssafetyhard$ are constraints that a repair cannot modify 
(see Sec.~\ref{sec:manager_and_repair}).
Our system's hard constraints
$\syssafetyhard$
only allow the robot to execute one skill at a time:
    $\square\big(\neg (\bigcirc\skill \land \bigcirc\skill')\big)$, for any two different skills $\skill, \skill' \in \out$. 
Given that, 
we encode the following hard assumptions $\envsafetyhard$:
\begin{enumerate*}[label=(\roman*),ref=\roman*]
    \item the uncontrollable terrain and request inputs 
    cannot change during execution:
    $\square (\var \leftrightarrow \bigcirc \var)$,
    $\forall \var \in \inpterrain \cup \inprequest$,
    and
    \item the robot inputs $\inpcen$ remain unchanged if no skill is executed: 
    $\square\big(\bigwedge_{\skill\in\out} \neg \skill \to \bigwedge_{\var\in\inpcen} (\var \leftrightarrow \bigcirc \var)\big)$.
\end{enumerate*}

 \subsection{High-level Manager and Symbolic Repair}\label{sec:manager_and_repair}
%

We design a high-level manager
to 
efficiently synthesize controllers for
the encoded specification
under given sets of terrain and request states.
%
%
The manager takes in 
the specification $\spec$,
a predefined set of terrain states 
$\inpstatesterrainexprposs \subseteq \inpstatesterrain$,
and
a predefined set of request states 
$\inpstatesrequestposs \subseteq \inpstatesrequest$.
For every pair of integer terrain and request state 
$(\inpstateterrain, \inpstaterequest) \in \inpstatesterrainexprposs \times \inpstatesrequestposs$,
the manager 
first translates
the integer terrain state $\inpstateterrain: \inpterrain \to \range{n_t}$
to its corresponding Boolean terrain state
$\inpstateterrainbool: \inpterrainbool \to \{\true, \false\}$,
using a set of Boolean propositions $\inpterrainbool$ to represent the integer terrain inputs $\inpterrain$,
as done in~\cite{ehlers2016slugs}. 
We overload $\inpstateterrainbool$ and $\inpstaterequest$
to represent the inputs that are $\true$
in $\inpstateterrainbool$ and $\inpstaterequest$.
The manager then
generates a partial evaluation
$\spec' \coloneqq \spec[\inpstateterrainbool \cup \inpstaterequest, \inpterrainbool \cup \inprequest \setminus \inpstateterrainbool \cup \inpstaterequest]$ (see Definition~\ref{def:partial_eval}).
Next,
we remove skills that cannot be executed in the local environment, i.e., their preconditions are evaluated to $\false$ under the partial evaluation.
Since
the inputs of $\spec'$ only include the robot inputs $\inpcen$
and the outputs of $\spec'$ consist of fewer skills,
we avoid the exponential blow-up of the synthesis algorithm in the size of the variables.
Lastly,
the manager synthesizes a strategy $\strategy_{\spec'}$ for each $\spec'$.

The partial evaluation $\spec'$ can be unrealizable
if the gait-fixed MICP-based 
physical feasibility checking 
in Sec.~\ref{sec:motion_primitive} 
does not generate enough skills for the robot to reach the requested goal under the terrain state.
To add more robot skills,
we will leverage a gait-free MICP 
which retains all continuous decision variables and constraints from the gait-fixed MICP (formulated in Fig.~\ref{fig:micp}) 
but modifies the contact state-dependent constraints 
so that the contact sequence and timing are no longer fixed.
Since
such a gait-free MICP is 
computationally expensive,
we leverage a symbolic repair tool~\cite{pacheck2022physically}
to 
generate symbolic suggestions that 
guide us to perform 
necessary gait-free MICP only.
Given an unrealizable $\spec'$,
symbolic repair systematically
modifies the pre- and postconditions of existing skills
to suggest new skills that make $\spec'$ 
realizable,
if such skills exist and are physically feasible.
Fig.~\ref{fig:preliminary_nine_squares}(c) illustrates the use of repair in Example~\ref{example:1}. 
To create a skill that reaches the requested grid $(x_0, y_0)$,
repair selects the skill transition that moves the robot from the grid $(x_1, y_0)$ to $(x_2, y_0)$,
and modifies the postcondition to be $(x_0, y_0)$.

\section{Online Execution}\label{sec:online_execution}
The online execution module takes as input a set of online polygons $\mathcal{P}_{\text{on}}$ and corresponding integer terrain state $\inpstateterrain \in \inpstatesterrain$,
an online local goal and its corresponding request state $\inpstaterequest \in \inpstatesrequest$,
and
a set of strategies $\strategy = \{\strategy_{\spec'}\}$
from the offline synthesis module (Sec.~\ref{sec:reactive_gait_synthesis}).
As shown in Fig.~\ref{fig:combined_offline_online}(c),
if the robot encounters an unseen terrain or request state,
i.e. $(\inpstateterrain, \inpstaterequest) \not\in \inpstatesterrainexprposs \times \inpstatesrequestposs$,
or fails to execute a symbolic transition,
we leverage runtime repair to
create new robot skills and 
synthesize a new strategy $\strategy_{\spec'}$
(Sec.~\ref{sec:online_repair}).
%
Next, we execute the strategy $\strategy_{\spec'}$
corresponding to $(\inpstateterrain, \inpstaterequest)$.
We execute each transition in $\strategy_{\spec'}$
by solving a MICP 
with more detailed and accurate terrain information 
to generate a nominal trajectory
(Sec.~\ref{sec:automaton_online_evaluation}). 
We use a tracking controller to follow the nominal trajectory in real time.
%
(Sec.~\ref{sec:tracking_control}).
After reaching the request state,
the robot resets the strategy with new terrain and request states,
and repeats the execution
until reaching the final goal state.


\subsection{Runtime Repair}\label{sec:online_repair}
The runtime repair takes in
a terrain state $\inpstateterrain \in \inpstatesterrain$, 
a request state $\inpstaterequest \in \inpstatesrequest$,
and a Boolean formula $\disallowedtrans$
representing disallowed transitions discovered at runtime.
We first update the system hard constraint $\syssafety$ in the specification $\spec$ from Sec.~\ref{sec:spec_gen}
to be $\syssafety \land \square \disallowedtrans$,
so that the updated specification respects the newly discovered disallowed transitions, if any.
We then
leverage the high-level manager to create a partial evaluation of $\spec$ over $\inpstateterrain$ and $\inpstaterequest$,
and perform symbolic repair 
to generate new robot skills and a new strategy to handle $\inpstateterrain$ and $\inpstaterequest$,
as done in Sec.~\ref{sec:manager_and_repair}.

\subsection{Strategy Execution}\label{sec:automaton_online_evaluation}
The red path in Fig.~\ref{fig:combined_offline_online}(c) represents the 
automaton execution process. 
We first solve an online MICP before transitioning to a new symbolic state.
The automaton advances only if the MICP successfully finds a solution.
Slightly different from the offline gait-fixed MICP formulation in Eq.~\ref{ocp:mip_general},
the online MICP takes in online polygons $\mathcal{P}_{\text{on}}$ from a terrain segmentation module,
the chosen gait, 
and initial and final robot states from the symbolic transition.
If the MICP is not successful,
we set the symbolic transition as a disallowed transition and perform runtime repair (Sec.~\ref{sec:online_repair}).



\subsection{Tracking Control}\label{sec:tracking_control}
The tracking control module adopts an MPC-Whole-Body-Control (WBC) hierarchy, ensuring precise end-effector (EE) and centroidal momentum tracking. The MPC tracks the reference trajectory produced by the online MICP using a more accurate model of the robot's centroidal dynamics and kinematics, and is solved via the OCS2 library \cite{OCS2}. The state estimator fuses IMU data, joint encoders, and motion capture inputs to provide accurate body position information.


\section{Results}\label{sec:experiments}
We present examples of maneuvering across diverse environments in a Gazebo simulation to demonstrate the framework's efficacy, scalability, and generalizability. Case studies on online execution highlight our framework's capability to handle unforeseen terrains and failures. 


\begin{figure}[t!]
    \centering
    \includegraphics[width=\linewidth]{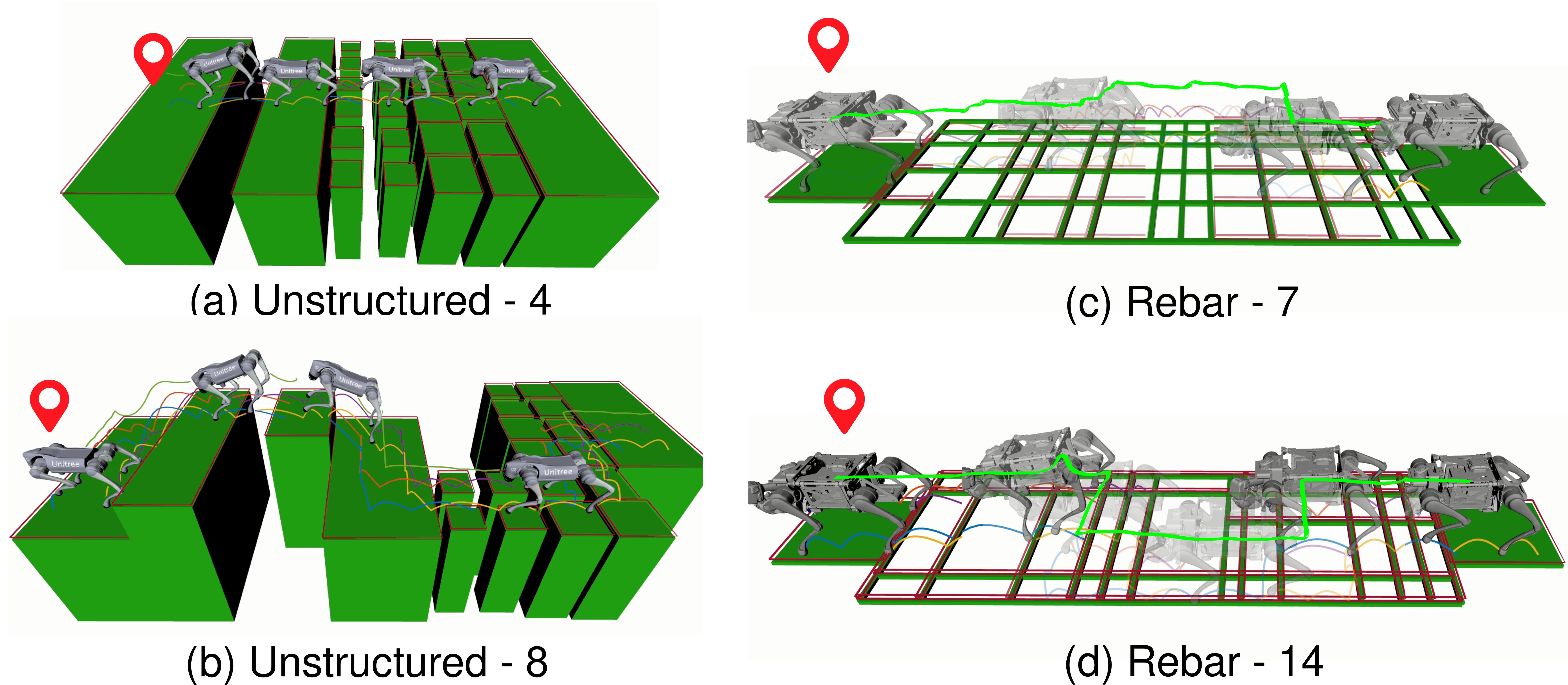}
    \caption{Unstructured and rebar terrain scenarios.}
    \label{fig:online_terrains}
    \vspace{-0.2in}
\end{figure}

\begin{table*}[t]
\scriptsize
    \caption{Offline Synthesis and Repair Time}
    \centering
     \begin{tabular}{|c | c | c | c | c | c|}
         \hline
         Scenario & Grid Size & 
         \begin{tabular}{@{}c@{}}Terrain and Request State Pairs\\ (Success/Total)\end{tabular} 
         & \begin{tabular}{@{}c@{}}Number of Skills \\ (Original/New/Total)\end{tabular} & Symbolic Repair Time (s) & \begin{tabular}{@{}c@{}}Feasibility Check Time (s)\\ (Gait-Fixed/Gait-Free)\end{tabular}\\\hline
         Unstructured - 4 & $3 \times 3$ & 169/169 & 18/\textcolor{red}{13}/64 & 8.07 &  18.97/449.94 \\
         & $5 \times 5$ & 129/158 & 19/\textcolor{red}{9}/64 & 77.16 &  20.26/241.42 \\\hline
         Unstructured - 8 & $3 \times 3$ & 250/264 & 19/\textcolor{red}{20}/256 & 14.90 & 22.92/382.63 \\
         & $5 \times 5$ & 179/246 & 20/\textcolor{red}{16}/256 & 93.21 & 18.59/196.88\\\hline
         Rebar - 7& $3 \times 3$ & 
         196/196 & 18/\textcolor{red}{11}/196 & 7.44 &  15.63/417.55
         \\
         & $5 \times 5$ & 196/196 & 18/\textcolor{red}{10}/196 & 21.63 &  14.83/374.89 \\\hline
         Rebar - 14 & $3 \times 3$ & 237/237 & 20/\textcolor{red}{29}/784 & 10.34 & 21.92/701.38\\
         & $5 \times 5$ & 196/196 & 20/\textcolor{red}{18}/784 & 24.31 & 23.06/453.94 \\\hline
    \end{tabular}
    \label{tab:offline_synthesis_time}
    \vspace{-0.15in}
\end{table*}

\subsection{Setup}
To demonstrate the generalizability of the proposed framework, 
we evaluate it on two scenarios with 
varying terrain types, 
safe stepping regions,
and robot platforms--Unitree Go2 and 
SkyMul Chotu 
(a modified Unitree Go1 robot dedicated for rebar tying tasks). 
We
model
obstacles
as a distinct terrain type,
with
obstacle avoidance 
encoded as hard constraints in $\syssafety$, 
and the requested waypoint 
are assumed not to be obstacles.
We test both $3 \times 3$ and $5 \times 5$ grid abstractions.
The $5 \times 5$ grid has a 
longer planning horizon and 
greater decision complexity.
We use a 0.8 m cell size for the unstructured terrain and 0.6 m for the rebar,
which can be adjusted for practical applications.

\subsubsection{Unstructured Terrain}
We abstract 8 terrain types--\textit{flat terrain}, \textit{high terrain}, \textit{low terrain}, \textit{dense stone}, \textit{sparse stone}, \textit{gap}, \textit{high gap}, and \textit{low gap}--with 
classification criteria 
based on polygon heights, 
counts,
and 
areas relative to the abstracted cells. Note that a terrain is classified as a \textit{gap} when the ratio of its overlapping area with the abstracted cell falls below a threshold.
We define two terrain configurations,
one with four selected terrain types 
and another with all eight terrain types,
as shown in Figs.~\ref{fig:online_terrains}(a) and~\ref{fig:online_terrains}(b).

\subsubsection{Rebar Terrain}
Inspired by automating labor-intensive rebar tying tasks on construction sites \cite{asselmeier2024hierarchical}, we explore a second case study where the quadrupedal robot navigates a rebar mat. 
We model each rebar as a rectangular polygon with a 3 cm width. 
Unlike the stepping stone scenario, this setup offers a larger number of potential stepping polygons due to the higher rebar density. 
Observing that the robot's traversability depends on rebar sparsity in different directions—whether aligned with the robot's facing direction or not, 
within a given tolerance—we abstract and classify rebar types based on a combination of horizontal sparsity (perpendicular to the robot's facing direction) and vertical sparsity (parallel to it). 
By calculating the number and relative spacing of the rebars inside each abstracted cell, we categorize each direction's sparsity as  \textit{dense} (0.05 - 0.15 m), \textit{sparse} (0.15 - 0.35 m), or \textit{extreme sparse} (above 0.35 m), \textit{single}, \textit{none}. 
To reduce the complexity of the problem, we remove some challenging combinations such as \textit{none} or \textit{single} in both directions. As a result, we define two example configurations with 7 and 14 rebar terrain types. The 7-type scenario in Fig.~\ref{fig:online_terrains}(c) generates random rebar sparsity between 0.15 and 0.35 m, excluding the \textit{extreme sparse} type. In contrast, the 14-type scenario in Fig.~\ref{fig:online_terrains}(d)  includes rebar sparsity from 0.15 to 0.6 m, potentially covering more challenging combinations including the \textit{extreme sparse} ones.


\begin{figure}[t]
    \centering
    \includegraphics[width=0.9\linewidth]{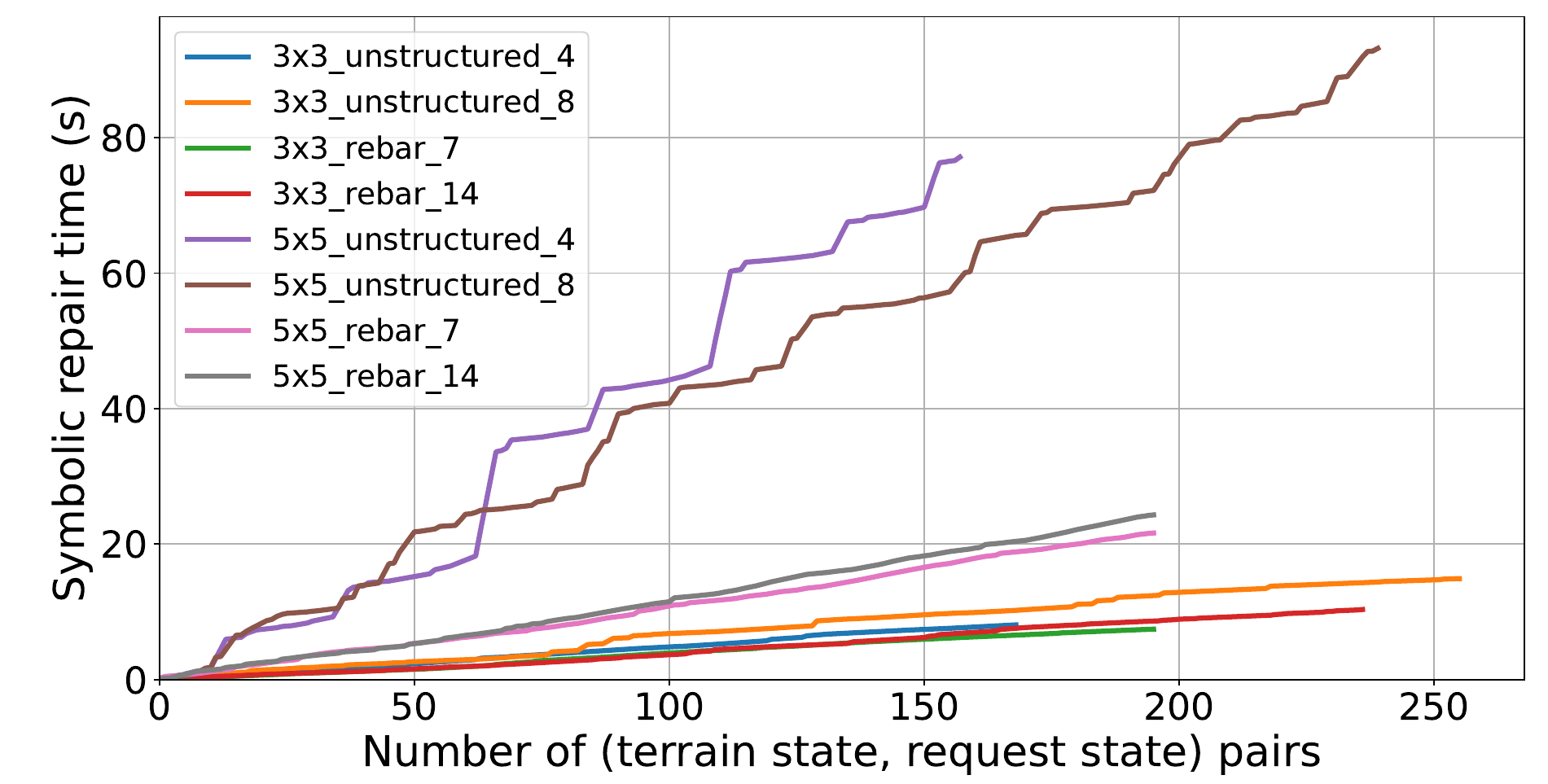}
    \caption{Symbolic repair time in the size of terrain and request state pairs.}
    \label{fig:repair_vs_num_states}
    \vspace{-0.15in}
\end{figure}

\subsection{Offline Synthesis Results}
We evaluate the offline synthesis module
for both \textit{Unstructured} and \textit{Rebar Terrain}
by initializing the skills with two trotting gaits ($2$ s and $3$ s).
Before synthesis, 
we generate the set of possible terrain states 
$\inpstatesterrainexprposs$ (supposedly provided by the user)
by systematically sweeping a local grid across the entire terrain map based on the best available knowledge.
The request states $\inpstatesrequestposs$ are defined as the top row of the local grid map in the robot's facing direction, along with the two corner cells to allow
additional movement directions. 

Table~\ref{tab:offline_synthesis_time} reports 
\begin{enumerate*}[label=(\roman*),ref=\roman*]
    \item the number of terrain and request states pairs $(\inpstateterrain, \inpstaterequest) \in \inpstatesterrainexprposs \times \inpstatesrequestposs$, including both successful and total ones,
    \item the number of skills, 
    including the original,
    the newly discovered,
    and the total possible ones,
    and
    \item offline repair and feasibility checking times, including both from gait-fixed and gait-free MICP.
\end{enumerate*}
We first note that
repair solves
$93.4\%$ of the terrain and request state pairs. 
The rest are unrepairable because the request states are unreachable due to obstacles or physical infeasibility.
As highlighted in red,
the number of newly discovered skills 
is $71.6 - 97.6\%$ 
smaller than checking all possible skills,
each of which 
corresponds to solving an expensive gait-free MICP
that takes $27.23$ s on average (max $145.66$ s) for a new locomotion gait.
This demonstrates that offline repair effectively minimizes the number of gait-free MICP solves.

To investigate the scalability of offline repair,
we perform repair 
across various sizes of the terrain and request states.
As shown in
Fig.~\ref{fig:repair_vs_num_states},
the repair runtime grows linearly in the size of the terrain and request states for both $3\times 3$ and $5 \times 5$ grids.
While fewer terrain and request states handled offline reduce checked time, they may lead to more frequent unforeseen states during online execution.
Moreover, we note that
the slope of $5 \times 5$ cases in Fig.~\ref{fig:repair_vs_num_states}
are steeper than those of $3\times 3$
as a result of the increased state space.
On average,
repair takes 478.1$\%$ more time for $5\times 5$ grids than $3\times 3$ for one terrain and request states pair.
%
In practice, 
the perception range and the robot size limit the grid size, 
so we do not test beyond $5\times 5$ grids,
though the user can adjust the resolution.



\begin{figure}[t]
    \centering
    \includegraphics[width=\linewidth]{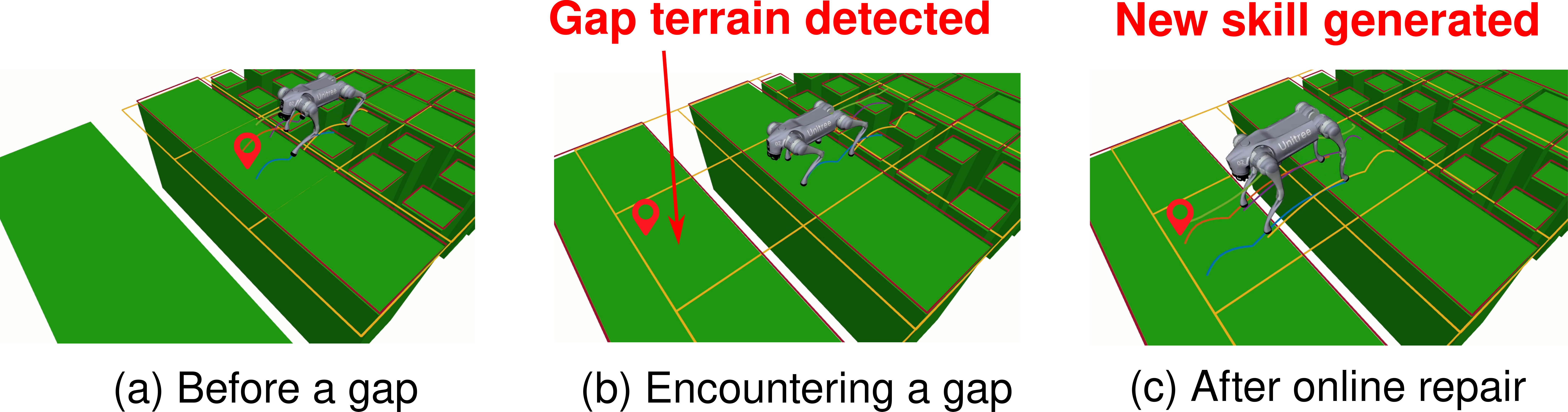}
    \caption{Runtime repair scenario when encountering a gap.}
    \label{fig:online_repair}
    \vspace{-0.15in}
\end{figure}

\subsection{Online Execution Results}
Fig.~\ref{fig:online_terrains} visualizes the snapshots of Go2 and Chotu traversing the terrains during online execution. The robot is tasked with reaching a global waypoint using a naive global shortest path planner. 
Specifically, the local waypoint in the local grid map is selected by choosing the closest, non-obstacle grid cell toward the global waypoint.
Fig.~\ref{fig:online_terrains}(a) shows a maneuver traversing through flat terrain, dense stepping stone, sparse stepping stone, and a gap. Fig.~\ref{fig:online_terrains}(b) shows the robot navigate through all eight terrain types, including elevation variations by adaptively choosing the locomotion gaits. Similarly, Fig.~\ref{fig:online_terrains}(c) and (d) present rebar traversing in separate scenarios with potentially 7 and 14 rebar terrain types. Notably, the robot leverages a set of newly discovered leaping gaits with short aerial phases for transitions from lower to higher terrain or across gaps and extreme sparse rebars as shown in Fig.~\ref{fig:online_terrains}(b) and (d).
On average, the online gait-fixed MICP takes $430$ ms to solve the unstructured terrain cases and $1.1$ s for rebar cases when the collision avoidance is disabled. 
The rebar scenarios exhibit $155.8\%$ longer solve times due to their overlapping, skewed rebar arrangement, and a larger number of terrain polygons. 
Additionally, 
enabling 
collision avoidance (necessary for \textit{Unstructured - 8} case to handle the elevation change) 
increases the number of binary variables
by $271.0\%$, 
increasing the solve time to $2.65$ s.

As a case study to demonstrate the capability of handling unforeseen terrains and failures during execution, 
we highlight the following run-time scenarios:
\subsubsection{Unforeseen Terrain States} 
With limited knowledge of the global terrain polygons, the terrain states provided during offline synthesis may be insufficient when encountering new terrain configurations. Fig.~\ref{fig:online_repair} illustrates this in an Unstructured - 4 types scenario with a $3 \times 3$ grid size, where terrain states involving \textit{gap} terrain are excluded, treating \textit{gap} as an unseen terrain type during offline synthesis. When the robot reaches a new terrain state that includes a \textit{gap}, we trigger runtime repair for the current terrain and request states. A new skill with a leaping gait is generated to successfully cross the gap. 
The runtime repair takes $12.36$ s, where symbolic repair 
takes $0.18$ s, and the gait-free MICP takes $12.18$ s to generate the new leaping gait.

\subsubsection{Solving Failure} Another common runtime failure arises from MICP solving failures. Due to discrepancies between the predefined polygons used in offline synthesis and the actual online terrains—such as variations in shape and size—a skill deemed feasible offline may become infeasible when executed online, even if classified under the same symbolic transition. 
%
Originally moving straight ahead, the detours shown as green lines in Fig.~\ref{fig:online_terrains}(c) and (d) illustrate the re-synthesis process triggered by solving failures in both the 7-type and 14-type cases.
Solving failure occurs more frequently in the rebar scenario than in the unstructured terrain scenario due to uncertainties in the shapes and sizes of online rebar polygons. 
The runtime resynthesis takes $0.11$ s and $0.07$ s 
for the two cases
respectively. 
Both cases do not trigger runtime repair 
because other existing skills suffice to navigate the robot to the goals under the solving failures.

\section{Conclusion}
In this work, we presented an integrated planning framework for terrain-adaptive locomotion that combines reactive synthesis with MICP, enabling safe and reactive responses in dynamically changing environments. 
Our approach not only enhances motion feasibility and safety but also provides a scalable solution for legged robots maneuvering in complex and safety-critical environments. 
Future work will involve extensive hardware testing to validate the proposed framework 
and generalization to heterogeneous robot teaming \cite{zhou2022reactive}. 
Additionally, we plan to refine our symbolic repair approach to relieve the computational burden further
via
learning-based methods 
such as large language models.
%

\bibliographystyle{IEEEtran}
\bibliography{references}

\end{document}